\documentclass[conference]{IEEEtran}
\IEEEoverridecommandlockouts
\usepackage{cite}
\usepackage{amsmath,amssymb,amsfonts}
\usepackage{algorithmic}
\usepackage{graphicx}
\usepackage{textcomp}
\usepackage{xcolor}

\usepackage{hyperref}
\usepackage{booktabs}
\usepackage{multirow}
\usepackage{xcolor}
\usepackage{array}
\usepackage{adjustbox}
\usepackage{subcaption}
\usepackage{subcaption}
\usepackage{xcolor}
\usepackage{amssymb}

\newcommand{\upone}{\textcolor{green!60!black}{\scriptsize$\uparrow$}}
\newcommand{\uptwo}{\textcolor{green!60!black}{\scriptsize$\uparrow\uparrow$}}
\newcommand{\upthree}{\textcolor{green!60!black}{\scriptsize$\uparrow\uparrow\uparrow$}}

\newcommand{\downone}{\textcolor{red}{\scriptsize$\downarrow$}}
\newcommand{\downtwo}{\textcolor{red}{\scriptsize$\downarrow\downarrow$}}
\newcommand{\downthree}{\textcolor{red}{\scriptsize$\downarrow\downarrow\downarrow$}}

\def\BibTeX{{\rm B\kern-.05em{\sc i\kern-.025em b}\kern-.08em
    T\kern-.1667em\lower.7ex\hbox{E}\kern-.125emX}}
\begin{document}

\title{\LARGE \bf Why Domain Matters: A Preliminary Study of \\Domain Effects in Underwater Object Detection}

\author{
\IEEEauthorblockN{
Melanie Wille \qquad
Dimity Miller \qquad
Tobias Fischer \qquad 
Scarlett Raine
}
\IEEEauthorblockA{
\textit{QUT Centre for Robotics, Queensland University of Technology} \\
Brisbane, Australia
}
\IEEEauthorblockA{
\{willemc, d24.miller, tobias.fischer, sg.raine\}@qut.edu.au
}
}

\maketitle

\begin{abstract}
Domain shift, where deviations between training and deployment data distributions degrade model performance, is a key challenge in underwater environments. Existing benchmarks testing performance for underwater domain shift simulate variability through synthetic style transfer. This fails to capture intrinsic scene factors such as visibility, illumination, scene composition, or acquisition factors, limiting analysis of real-world effects. We propose a labeling framework that defines underwater domains using measurable image, scene, and acquisition characteristics. Unlike prior benchmarks, it captures physically meaningful factors, enabling semantically consistent image grouping and supporting domain-specific evaluation of detection performance including failure analysis. We validate this on public datasets, showing systematic variations across domain factors and revealing hidden failure modes.

\end{abstract}


\section{Introduction}
Underwater object detection is a critical tool for marine scientists, allowing efficient analysis of large-scale seafloor imagery to monitor benthic indicator species. This is crucial for assessing ecosystem health and the impact of increasing human activity on those environments~\cite{doig2025training}. Traditional workflows rely on collecting and annotating large amounts of training data to achieve the desired model performance. However, this process is not only resource-intensive, but must often be repeated for new sites with different environmental characteristics or collection protocols~\cite{han2023see, walker2024domain}. This is due to \emph{domain shift}, where new environments or collection platforms introduce variations in lighting, turbidity, depth, scene appearance, and object characteristics, significantly limiting data reusability~\cite{han2023see, elmezain2025advancing, nabahirwa2025structured}. 

\begin{figure}
    \centering
    \includegraphics[width=\linewidth]{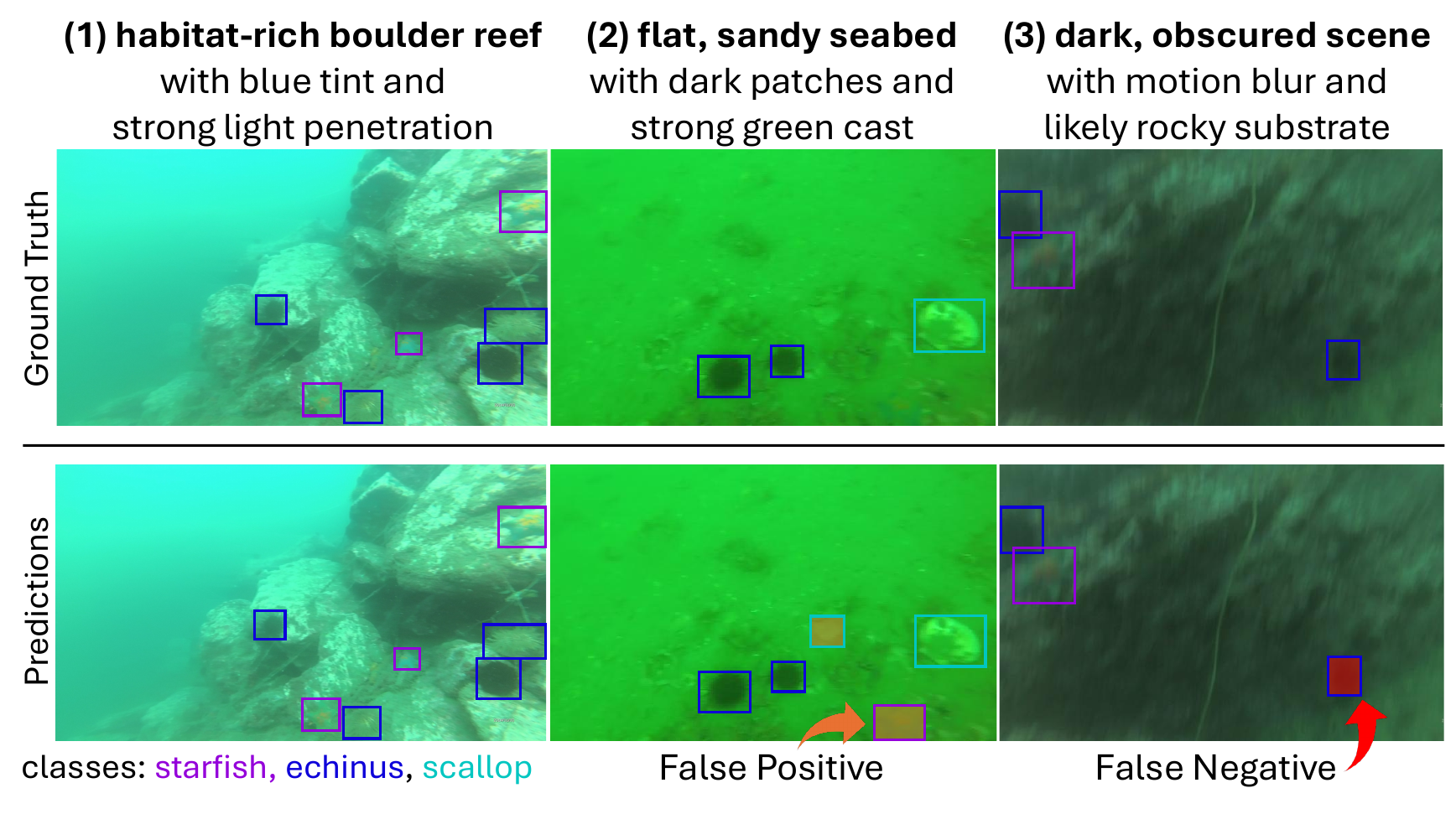}
    \vspace{-0.62cm}
    \caption{Detection performance across visually separable underwater domains (false positives indicated in orange and false negatives in red). Performance varies with conditions: \textit{echinus} stand out in well-lit images (columns 1, 2) but become difficult to distinguish from shadows in dark scenes (3); \textit{starfish}’s distinct shape can be separated well from textured rocks (1, 3) but confused in context-poor sand (2); algae-rich, green water reduces contrast, making objects harder to differentiate (2); motion blur degrades object boundaries (3). These examples highlight that detection accuracy is influenced by external factors tied to environmental and acquisition conditions, motivating our framework for domain-aware evaluation.
}
\vspace{-0.5cm}
    \label{fig:front_page}
\end{figure}

Fig.~\ref{fig:front_page} illustrates the domain shift effect in underwater object detection: the same model shows substantial variations in detection performance in different underwater environments. The same species may appear different or become more difficult to detect in a new location with changed conditions. Domain shift is therefore a fundamental challenge in underwater object detection~\cite{walker2024domain, elmezain2025advancing, liu2020towards, chen2023achieving}. 

While some underwater object detection datasets are collected under controlled acquisition at a defined location~\cite{ahmed2024improving, pedersen2019brackish}, many widely-used benchmarks mix data of diverse conditions from various sources~\cite{liu2021duo, fu2023ruod, jiang2021uodd}. Evaluating models on such data, especially using aggregate metrics like mAP over the entire dataset, can hide domain-specific failure modes and prevent a systematic analysis of how environmental and acquisition factors affect detection performance. A few works group images into different types 
to study domain generalization explicitly~\cite{liu2020towards, chen2023achieving}, but they use synthetic style transfer to simulate these image domains. While this approach changes low-level appearance, it does not consider the naturally correlated physical and semantic factors that play a crucial role in real-world scenarios, making these ``domains'' difficult to interpret and only weakly comparable to actual deployment conditions.

In this paper, we propose to explicitly characterize underwater domains using measurable, intrinsic properties of the image, scene, and data acquisition process. Rather than relying on synthetic or implicit domain definitions, we represent domain variability through physically meaningful factors. Concretely, we assign each image a set of categorical labels describing properties such as visibility, object layout, and camera perspective. These labels define domains as combinations of these properties, enabling a structured and interpretable analysis of domain-dependent detection performance.

To guide this study, we formulate two research questions:

\begin{itemize}
    \item \textbf{RQ1:} How can domain variability in underwater imagery be decomposed into interpretable and measurable factors that enable consistent grouping of images?

    \item \textbf{RQ2:} To what extent do domain-specific factors influence object detection performance across different conditions?
\end{itemize}

In answering these research questions, we make the following contributions:

\begin{enumerate}

    \item We introduce a framework to systematically characterize underwater domain shift by assigning each image a set of interpretable categorical labels 
    describing image appearance, scene composition, and acquisition geometry. These labels define domains as combinations of physically meaningful factors. 

    \item We apply this framework to introduce domain annotations for two public underwater object detection datasets, enabling structured analysis of how different conditions are represented in existing benchmarks.

    \item We provide empirical evidence that detection performance varies substantially across our defined domain categories, revealing systematic and sometimes counter-intuitive failure modes that are hidden by standard evaluation with aggregate metrics.

\end{enumerate}


\section{Related Work}
\label{sec:rel_work}
\subsection{Domain Shift in Underwater Object Detection}
Domain shift refers to changes in the input data distribution between training and deployment scenarios, resulting in performance degradation of computer vision models~\cite{walker2024domain}. This effect is worsened by the dynamic nature of underwater environments. Image properties vary significantly across locations, seasons, weather, and depth, due to differences in turbidity, light scattering and absorption, contrast, color, plankton presence, and temperature, which in turn influence marine species characteristics and seafloor structure~\cite{walker2024domain, niu2025domain, elmezain2025advancing}. These environmental factors cause domain shift on the overall image-level as well as instance-level for marine targets~\cite{han2023see}.

Recent work on sea urchin detection across multiple geographic locations~\cite{rawlinson2025urchinbot} highlighted data acquisition as a critical factor for underwater domain shift, attributing under-performance in one specific location to increased image blur and distance resulting from the data collection platform. These findings are based only on manual inspection of failure cases, underscoring the lack of structured approaches to understand domain-dependent performance.

\subsection{Approaches to Domain Shift}
Most works~\cite{wen2026joint, ren2026underwater, sun2025efcwm, luo2025hyuod, niu2025domain, xie2024fiod, saoud2024mars} adopt model-centric approaches to address domain shift by improving detector robustness to varying conditions. A common approach is domain adaptation through adversarial training with enhanced images~\cite{wen2026joint} or transfer learning leveraging limited samples for domain-specific fine-tuning~\cite{ren2026underwater, sun2025efcwm}. Others incorporate physics-based priors or frequency-domain representations to preserve feature information~\cite{luo2025hyuod, niu2025domain, xie2024fiod}. Additionally, architectural enhancements such as attention mechanisms and adaptive feature correction modules have been proposed to improve generalization~\cite{saoud2024mars, sun2025efcwm}. Although these approaches have made progress in improving general performance of models, analyzing the domain specific impacts on performance will enable us to develop tailored solutions to target specific conditions that will overall lead to improved performance across a range of conditions. From a data-based point of view, two datasets called S-URPC2019~\cite{liu2020towards} and S-UTDAC2020~\cite{chen2023achieving} have been proposed to explicitly study domain generalization in underwater environments. While existing domain generalization datasets for underwater object detection provide a useful starting point, their domain definitions are based on synthetic style transfer and are not tied to intrinsic physical or semantic properties. As a result, domain assignments may not reflect underlying environmental conditions, and images with differing characteristics can be grouped within one domain.

\subsection{Research Gap}
The limitations of existing underwater domain generalization datasets described above motivate new domain definitions grounded in intrinsic image, scene, and acquisition characteristics to enable semantically and physically meaningful grouping for analysis and evaluation. Several metrics have been proposed to quantify underwater image quality~\cite{panetta2015human, wang2018imaging, yang2021reference, yang2015uciqe}, but are primarily used to evaluate image enhancement and approximate human perception, not leveraged to analyze detector performance. This reveals a gap in current approaches, which lack interpretable, unified domain definitions for systematic evaluation. Inspired by this, our work constructs underwater domains from quantifiable properties to analyze domain shift effects on detection performance and failure modes, informing mitigation strategies.
\section{Proposed Domain Labeling Framework}

\begin{figure*}
    \centering
    \includegraphics[width=0.95\linewidth, trim=0.74cm 1.73cm 0.5cm 1.22cm, clip]{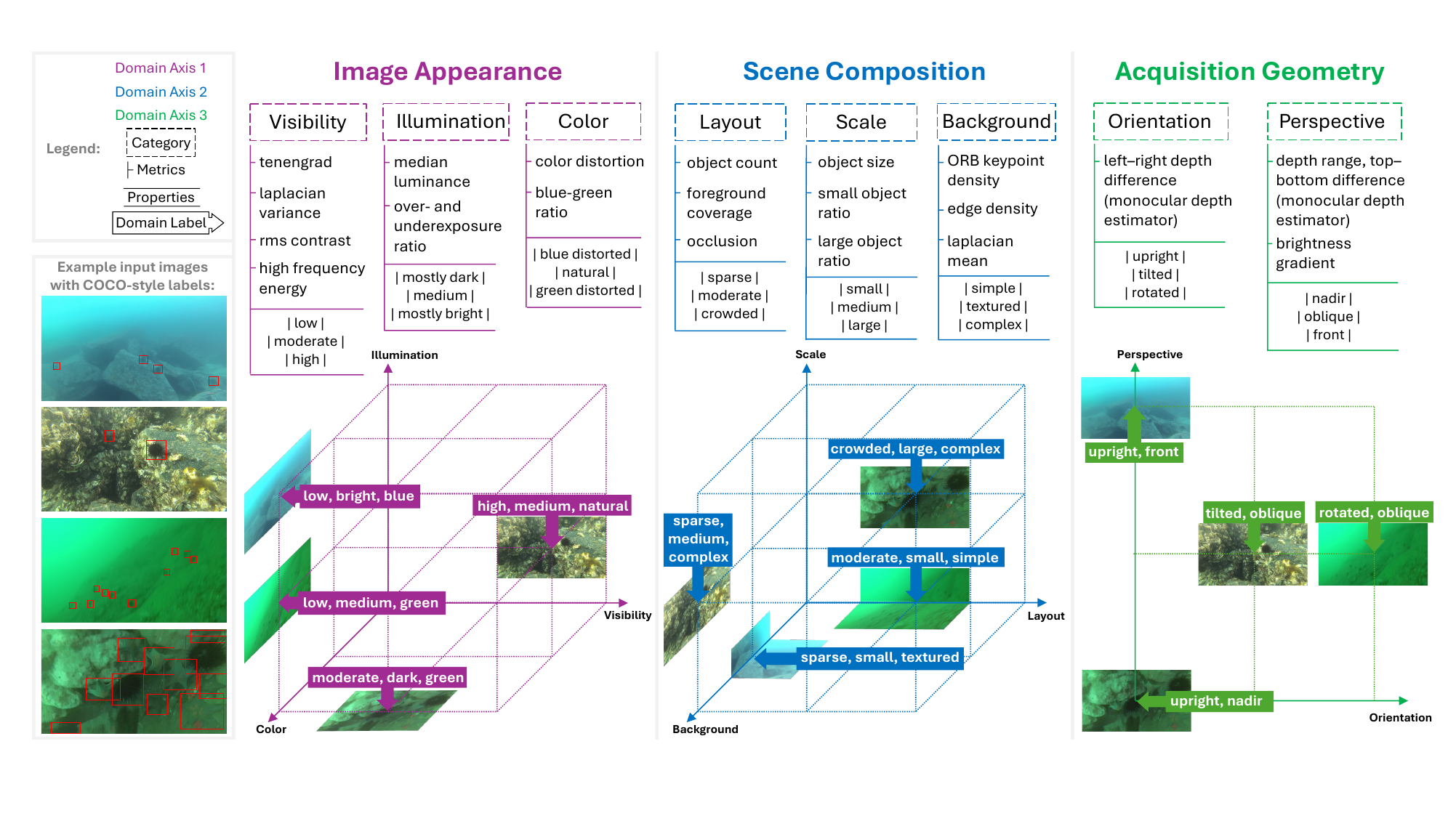}
    \caption{Overview of our underwater domain labeling framework. Images are assigned domain labels along three axes: image appearance (left, purple), scene composition (middle, blue), and acquisition geometry (right, green). Each axis is decomposed into interpretable categories (sub-axes), which are independently characterized using specified metrics. Images are assigned a domain label that describes their properties across all categories (e.g. low visibility, medium illumination, green color). Bottom left: selection of example images that is mapped onto the domain axes.}
    \vspace{-0.5cm}
    \label{fig:domain_labeling_framework}
\end{figure*}

In this section, we introduce an underwater domain labeling framework (Fig.~\ref{fig:domain_labeling_framework}). Building on prior studies addressing domain shift in underwater object detection~\cite{han2023see, walker2024domain, rawlinson2025urchinbot} (Section~\ref{sec:rel_work}), we observe that domain shift arises from three complementary and distinct sources: (1) image appearance, which varies due to turbidity and light scattering; (2) scene composition, reflecting the spatial arrangement of targets and the complexity of specific habitats; and (3) acquisition geometry, determined by camera orientation and viewpoint. Therefore, we model these factors as orthogonal axes, which are further decomposed into relevant categories and quantified using a combination of established image metrics and object-level properties, as described in the following sections.

\subsection{Metric Selection and Calibration}
To ensure that the selected metrics and thresholds are meaningful, we construct a manually annotated subset of 100 images sampled from four widely used underwater object detection datasets\footnote{Detecting Underwater Objects (DUO)~\cite{liu2021duo}, Rethinking general Underwater Object Detection (RUOD)~\cite{fu2023ruod}, Underwater Target Detection Algorithm Competition 2020~\cite{song2023boosting}, Underwater Object Detection Dataset~\cite{jiang2021uodd}}, chosen to cover the diverse domain properties. Each image is labeled across all domain categories by visual inspection and used to validate that candidate metrics exhibit consistent trends with the annotations (e.g. low-visibility images yield lower sharpness and contrast scores). The subset is used exclusively for calibration and complemented by further analysis of histogram statistics and value ranges of each metric across all data (not domain-labeled). We determine thresholds and score weightings based on distributional separation and agreement with manual labels. This iterative process is combined with visual validation on randomly sampled images to ensure semantic consistency of the assigned categories.

The final framework is implemented as an automated pipeline that derives domain labels based on the calibrated metrics and thresholds along each axis, described as follows.

\subsection{Axis 1: Image Appearance}

\textbf{Visibility.} 
Visibility reflects the overall clarity of image content, which is often degraded in underwater imagery due to scattering effects that reduce sharpness, contrast, and high-frequency detail. Underwater image quality metrics commonly assess sharpness using gradient-based edge operators such as Sobel filters~\cite{panetta2015human, yang2021reference}, which we approximate using the standard Tenengrad measure $T$. We complement this with the variance of the Laplacian $V_{\nabla^2}$, a widely used second-order measure for blur estimation. Contrast is another key component in underwater image quality metrics~\cite{panetta2015human, yang2015uciqe, yang2021reference}, which we quantify using RMS contrast $R$. Finally, since attenuation of high-frequency components reflects the loss of edges and textures~\cite{yang2021reference}, we include frequency-domain energy $F$. After log-scaling skewed distributions and clipped min-max normalization, these metrics are combined into a weighted visibility score $V = 0.35\,T + 0.30\,V_{\nabla^2} + 0.20\,R + 0.15\,F$, to categorize images as low $(V < 0.35)$, high $(V > 0.65)$, or moderate visibility (otherwise).

\textbf{Illumination.}
Illumination describes the overall brightness and exposure conditions of an image. In underwater image quality assessment, illumination is typically addressed through global intensity statistics, reflecting the influence of lighting conditions on perceived image quality~\cite{liu2023enhancing, yang2015uciqe, wang2018imaging}. Consistent with this, we approximate illumination using the median luminance $L_m = \mathrm{median}(I)$ of the grayscale image $I$, a robust primary measure of global brightness, indicating dark $(L_m < 100)$, bright $(L_m > 130)$, or medium-light images (otherwise). In addition to luminance, over- and under-exposure ratios are computed based on threshold values $L_{over} = 225$ and $L_{under} = 30$ as secondary criteria for extreme cases.

\textbf{Color.}
Wavelength-dependent light attenuation causes characteristic color distortion in underwater images, typically resulting in blue or green color casts. One of the fundamental underwater image quality metrics, UIQM~\cite{panetta2015human}, models this effect through channel deviation in the opponent RG–YB color space. We simplify this idea by directly measuring channel imbalance in RGB space using pairwise channel differences. Specifically, we define color distortion as $D = \sqrt{(\mu_R - \mu_G)^2 + (\mu_R - \mu_B)^2 + (\mu_G - \mu_B)^2}$, and additionally consider the blue-to-green ratio $\mathrm{BGR} = \frac{\mu_B}{\mu_G}$, with $\mu_R, \mu_G, \mu_B$ as the mean intensities of the red, green, and blue channels. Based on the results, images are grouped as blue distorted $(D > 0.6 \;\wedge\; \mathrm{BGR} > 0.8)$, green distorted $(D > 0.6 \;\wedge\; \mathrm{BGR} < 0.7)$, or natural (otherwise).

\subsection{Axis 2: Scene Composition}

\textbf{Layout.}
Layout describes the object density and spatial arrangement within the image using object count $N$, foreground\footnote{area covered by annotated objects} coverage $C = \frac{|\text{foreground pixels}|}{|\text{image}|}$
and overlap $O = \frac{\sum_{i \neq j} \text{area}(B_i \cap B_j)}{\sum_i \text{area}(B_i)}$, where $B_i$ denotes object bounding boxes, motivated by the established influence of occlusion on detection difficulty~\cite{nabahirwa2025structured}. We assign sparse $(N \leq 4 \;\wedge\; C < 0.05 \;\wedge\; O < 0.05)$, crowded $(N \geq 12 \;\vee\; C > 0.4 \;\vee\; O > 0.15)$, or moderate layout labels (otherwise).

\textbf{Scale.}
Scale describes the relative size of objects in the image which is often studied in small object detection works as challenging detection task~\cite{elmezain2025advancing}. We compute the mean normalized object area $A = \frac{1}{N} \sum_i \frac{\text{area}(B_i)}{|\text{image}|}$, as well as the ratio of small ($R_{small}$ for $A_{small} < 0.005$) and large ($R_{large}$ for $A_{large} > 0.025$) objects to distinguish between images dominated by small $(R_{\text{small}} \geq 0.5 \;\vee\; S < 0.005)$ or large $(R_{\text{large}} \geq 0.5 \;\vee\; S > 0.025)$ targets. Remaining images are considered medium scale.

\textbf{Background.}
Background considers the level of complexity and structural detail in non-object regions. Inspired by recent work that correlates edge density and feature-based measures to perceived visual complexity~\cite{chu2025makes}, we quantify our background category using keypoint density (ORB)~\cite{rublee2011orb} $K$, edge density~\cite{canny1986edge} $E$, and Laplacian mean $M_{\nabla^2}$, computed on background pixels only. After log-scaling and clipped min-max normalization, these metrics are combined into a background complexity score:
$B = 0.45\,K + 0.35\,E + 0.20\,M_{\nabla^2}$. Images are categorized as simple $(B < 0.15)$, complex $(B > 0.4)$, or textured (otherwise).

\subsection{Axis 3: Acquisition Geometry}
Viewpoint is known to impact object detection performance~\cite{wang2017viewpoint}. However, underwater imagery often lacks consistent acquisition metadata for this critical factor. As prior work has demonstrated that camera pose and scene structure can be inferred from depth map metrics~\cite{cin2023multi}, we estimate the following geometric properties using the state-of-the-art monocular depth estimator Depth Anything V2~\cite{depth_anything_v2}.

\textbf{Orientation.}
Orientation refers to the horizontal alignment of the camera relative to the scene. We estimate it using the depth difference between the left and right regions of the background $\Delta_{\text{lr}} = \left| \bar{D}_{\text{left}} - \bar{D}_{\text{right}} \right|$, where $\bar{D}$ denotes the mean depth. Orientation endpoints are upright $(\Delta_{\text{lr}} < 1)$ and rotated $(\Delta_{\text{lr}} > 2.5)$, while images between are slightly tilted.

\textbf{Perspective.}
Perspective captures the viewing angle of the camera relative to the scene. We rely on the vertical depth difference $\Delta_{\text{tb}} = \left| \bar{D}_{\text{top}} - \bar{D}_{\text{bottom}} \right|$, assuming that a nadir (top-down) view would lead to minimal depth variation across the scene, while front-view images are expected to show a larger depth range $R_D = \max(D) - \min(D)$. However, when the camera captures a lot of open water facing forward, there are few depth cues resulting in a flat depth map, which is why we include a brightness gradient $G_B = \bar{I}_{\text{top}} - \bar{I}_{\text{bottom}}$ as fallback. Based on these metrics, images are assigned either nadir $(\Delta_{\text{tb}} < 2 \;\wedge\; R_D < 3)$, front $(\Delta_{\text{tb}} > 4 \;\vee\; R_D > 5 \;\vee\; G_B > 50)$, or oblique (otherwise) perspective labels.

\section{Experimental Setup}
\begin{figure}
    \centering
    \includegraphics[width=\linewidth, trim=0.5cm 2cm 0cm 0.2cm, clip]{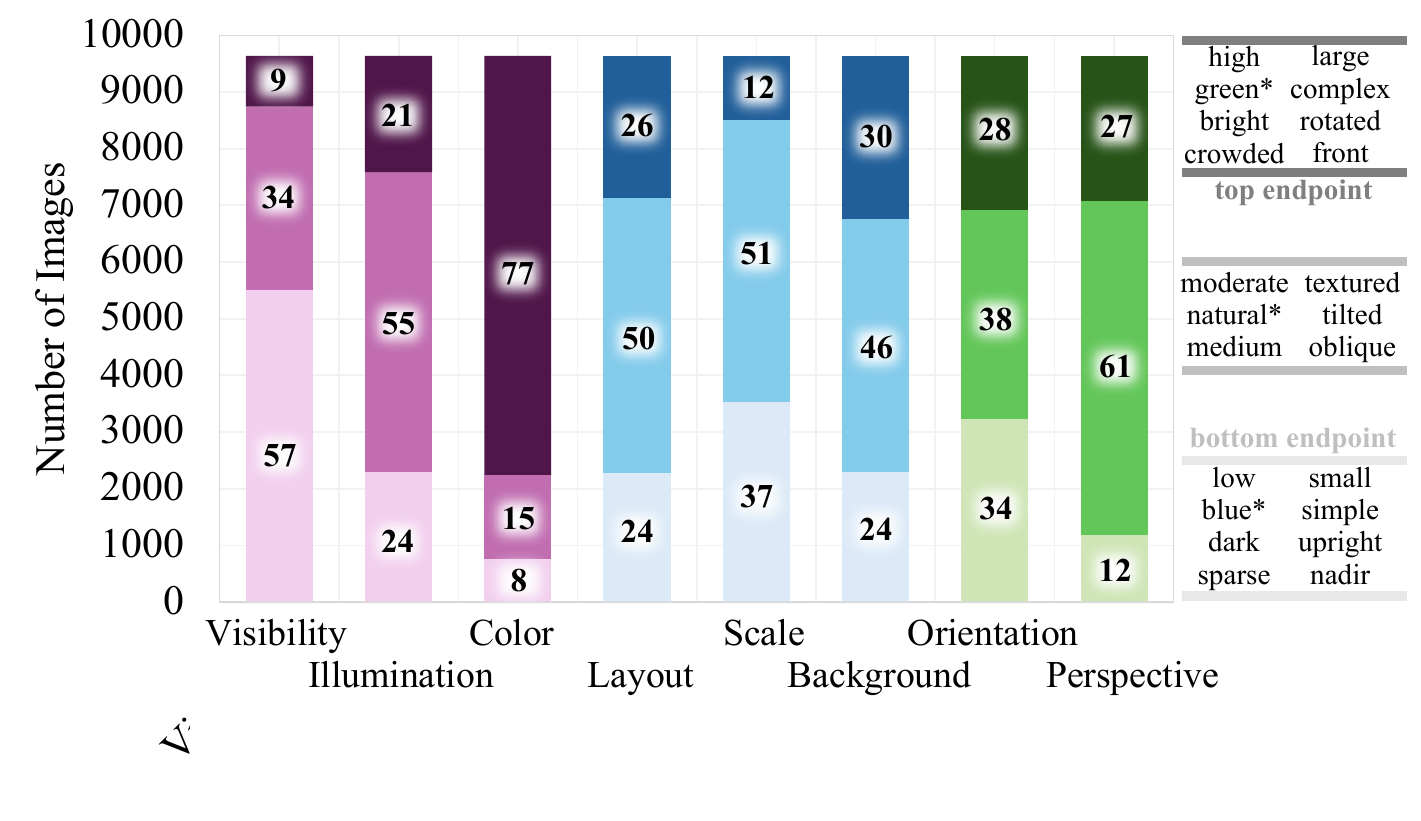}
    \vspace{-0.53cm}
    \caption{Distribution of images in the training set across all domain axes and categories. Colored bar segments represent category properties (see legend on the right), with numbers indicating proportions (\%). Evaluation is performed only on the bottom and top endpoints for all categories except color, which has non-ordered properties (indicated by *).}
    \vspace{-0.5cm}
    \label{fig:train_domain_distr}
\end{figure}

\textbf{Datasets.} We apply our proposed framework to assign domain labels to two public underwater object detection datasets: DUO~\cite{liu2021duo} and a four-class subset of RUOD~\cite{fu2023ruod}, introduced as RUOD-4C~\cite{wille2026all}. For higher statistical significance, we combine both subsets into one dataset for our experiments (12,050 images with 108,962 annotations across 4 classes) of which a random 80\% are assigned to the train split, 10\% for validation, and 10\% for testing. Fig.~\ref{fig:train_domain_distr} shows the resulting training distribution.

\textbf{Detection model.} We employ the most recent YOLO model, YOLO26n, from Ultralytics. The model is trained for 50 epochs using standard batch and image sizes (16, 640). We select the best-performing weights based on validation performance for independent evaluation on the test set.

\textbf{Evaluation protocol.} The model is trained on the full training split with mixed domains, but tested for each domain category separately. We only evaluate on the corresponding subset of images from the test set. For domains with inherent ordering (all but color), we focus on the end categories and exclude intermediate samples to test the extremes of domain shift. 
We use standard performance metrics precision $p = \frac{TP}{(TP + FP)}$, recall $r = \frac{TP}{(TP + FN)}$, and mean Average Precision $mAP = \frac{1}{C} \sum_{i=1}^{C} AP_i$, with $AP = \int_0^1 p(r) \, dr$.  We further perform failure analysis by computing raw false positives (FP) and false negatives (FN) per domain at IoU = 0.5 and a fixed confidence threshold of 0.5.  




\section{Results and Discussion}
We analyze detection performance across domain categories and complement standard metrics (Table~\ref{tab:condition_breakdown}) with error statistics (Fig.~\ref{fig:error_rates}). As described in the following sections, we observe consistent performance variations across all axes.

\begin{table*}
\centering
\caption{Performance across test-set domain conditions. Bold values indicate the best performance within each domain category. Arrows denote relative change with respect to the full mixed test set (which includes all domain properties, including the intermediate category omitted in extreme-case evaluation). Arrow direction and color indicate improvement (green, upward) or degradation (red, downward), while the number of arrows represents the magnitude of change: slight ($\uparrow$, 0--3\%), moderate ($\uparrow\uparrow$, 3--8\%), and strong ($\uparrow\uparrow\uparrow$, $>$8\%).}
\label{tab:condition_breakdown}
\renewcommand{\arraystretch}{1.15}
\setlength{\tabcolsep}{4pt}

\begin{adjustbox}{width=\textwidth}
\begin{tabular}{ll*{17}{c}}
\toprule
\multirow{3}{*}{\textbf{Metrics}} & \multirow{3}{*}{\textbf{Mixed}} 
& \multicolumn{7}{c}{\textbf{Image Appearance}}
& \multicolumn{6}{c}{\textbf{Scene Composition}}
& \multicolumn{4}{c}{\textbf{Acquisition Geometry}} \\
\cmidrule(lr){3-9} \cmidrule(lr){10-15} \cmidrule(lr){16-19}
& 
& \multicolumn{2}{c}{\textbf{Visibility}}
& \multicolumn{2}{c}{\textbf{Illumination}}
& \multicolumn{3}{c}{\textbf{Color}}
& \multicolumn{2}{c}{\textbf{Layout}}
& \multicolumn{2}{c}{\textbf{Scale}}
& \multicolumn{2}{c}{\textbf{Background}}
& \multicolumn{2}{c}{\textbf{Orientation}}
& \multicolumn{2}{c}{\textbf{Perspective}} \\
\cmidrule(lr){3-4} \cmidrule(lr){5-6} \cmidrule(lr){7-9}
\cmidrule(lr){10-11} \cmidrule(lr){12-13} \cmidrule(lr){14-15}
\cmidrule(lr){16-17} \cmidrule(lr){18-19}
& 
& Low & High & Dark & Bright & Blue & Natural & Green 
& Sparse & Crowded & Small & Large & Simple & Complex
& Upright & Rotated & Nadir & Front \\
\midrule
mAP50
& 0.868
& 0.836\,\downtwo
& \textbf{0.935}\,\uptwo
& 0.863\,\downone
& \textbf{0.901}\,\uptwo
& \textbf{0.927}\,\uptwo
& 0.837\,\downtwo
& 0.859\,\downone
& 0.796\,\downthree
& \textbf{0.873}\,\upone
& 0.835\,\downtwo
& \textbf{0.917}\,\uptwo
& 0.742\,\downthree
& \textbf{0.900}\,\uptwo
& \textbf{0.865}\,\downone
& 0.835\,\downtwo
& 0.845\,\downone
& \textbf{0.902}\,\uptwo \\

mAP50-95
& 0.649
& 0.627\,\downtwo
& \textbf{0.709}\,\upthree
& 0.627\,\downtwo
& \textbf{0.677}\,\uptwo
& \textbf{0.683}\,\uptwo
& 0.659\,\upone
& 0.638\,\downone
& 0.581\,\downthree
& \textbf{0.648}\,\downone
& 0.598\,\downtwo
& \textbf{0.715}\,\upthree
& 0.572\,\downthree
& \textbf{0.677}\,\uptwo
& \textbf{0.645}\,\downone
& 0.625\,\downtwo
& 0.633\,\downone
& \textbf{0.686}\,\uptwo \\

Precision
& 0.845
& 0.830\,\downone
& \textbf{0.930}\,\upthree
& 0.840\,\downone
& \textbf{0.877}\,\uptwo
& \textbf{0.899}\,\uptwo
& 0.861\,\upone
& 0.832\,\downone
& 0.762\,\downthree
& \textbf{0.849}\,\upone
& 0.829\,\downone
& \textbf{0.879}\,\uptwo
& 0.747\,\downthree
& \textbf{0.886}\,\uptwo
& \textbf{0.848}\,\upone
& 0.796\,\downtwo
& 0.855\,\upone
& \textbf{0.872}\,\uptwo \\

Recall
& 0.801
& 0.759\,\downtwo
& \textbf{0.875}\,\upthree
& 0.795\,\downone
& \textbf{0.839}\,\uptwo
& \textbf{0.880}\,\upthree
& 0.750\,\downtwo
& 0.794\,\downone
& 0.739\,\downtwo
& \textbf{0.813}\,\upone
& 0.747\,\downtwo
& \textbf{0.869}\,\upthree
& 0.670\,\downthree
& \textbf{0.838}\,\uptwo
& \textbf{0.791}\,\downone
& 0.794\,\downone
& 0.743\,\downtwo
& \textbf{0.838}\,\uptwo \\
\bottomrule
\end{tabular}
\end{adjustbox}
\vspace{-0.4cm}
\end{table*}

\subsection{Axis 1: Image Appearance}
\begin{figure}
    \centering
    \vspace{-0.2cm}
    \includegraphics[width=\linewidth]{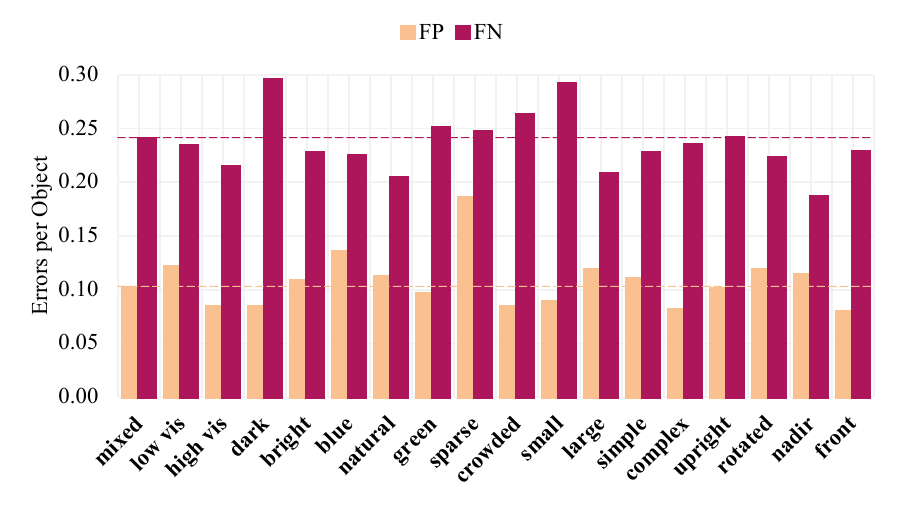}
    \vspace{-0.7cm}
    \caption{Average number of false positives (FP) and false negatives (FN) per object for each domain category at IoU = 0.5 and confidence threshold 0.5, revealing domain-dependent failure modes.}
    \vspace{-0.55cm}
    \label{fig:error_rates}
\end{figure}

\textbf{Visibility.} 
Visibility appears to be a major driver of domain shift: high visibility images strongly outperform low visibility across all performance metrics, e.g. by $\sim$10 points for mAP50 and precision and $\sim$12 points for recall (Table~\ref{tab:condition_breakdown}). Compared to the mixed test set, high visibility improves performance, while low visibility underperforms, especially in recall. This indicates missing objects is a major issue, confirmed by error rates (Fig.~\ref{fig:error_rates}), where images with low visibility incur more false negatives (FNs) and more false positive (FPs) per object. Thus, degraded visibility affects both detectability and reliability. This performance trend may be attributed to weaker object boundaries and missing fine details in low visibility, as blur and haze directly impact feature clarity. Overall, our visibility-based data split is meaningful
and discriminative, validating the relevance of our domain definition.

\textbf{Illumination.} 
Illumination influences performance more moderately. Bright conditions improve performance, while dark conditions show a slight but uniform performance drop, resulting in a 4--5 point gap between the extremes across all metrics (Table~\ref{tab:condition_breakdown}). Examining the distribution of errors (Fig.~\ref{fig:error_rates}), dark images are prone to missed detections, with the highest FN rate across all categories. Bright images exhibit lower FN and higher FP rates. This suggests that the model is more conservative in low-light conditions, likely because the reduced photon count increases noise and weakens signal strength. While illumination is a secondary domain factor compared to visibility, our framework still successfully shows it has an impact on performance.

\textbf{Color.} 
Color reveals a counter-intuitive mismatch between performance and training data: Table~\ref{tab:condition_breakdown} shows that blue images perform best across all metrics, notably above the mixed reference, despite being underrepresented in the training data (Fig.~\ref{fig:train_domain_distr}). On the other hand, green images exhibit similar performance to natural images. In terms of failure modes, green images show the highest FN rate, confirming detectability challenges, while blue images have fewer missed objects but more FPs, suggesting that more predictions are being made. We hypothesize that this is linked to environmental conditions: green water is often associated with phytoplankton and suspended particles, leading to increased scattering and reduced contrast, while blue water is often found in clearer and better illuminated open ocean sites. This is supported by dataset statistics, where blue images are disproportionately associated with high visibility (61\% of blue images vs 9\% of mixed images) and brightness (60\% of blue images vs 21\% of mixed images). This highlights that our domain labels capture meaningful physical differences and reveal hidden difficulty levels.

\subsection{Axis 2: Scene Composition}

\textbf{Layout.} 
Layout reveals a strong but counter-intuitive effect on performance. Contrary to the expectation that crowded scenes are more difficult (due to increased occlusion for example), crowded images consistently outperform sparse images across all metrics, with gaps of $\sim$8 points for mAP50, precision, and recall, and $\sim$6 points for mAP50-95 (Table~\ref{tab:condition_breakdown}). While crowded scenes slightly exceed the mixed baseline, sparse scenes show one of the largest performance drops across all domains. Notably, error rates (Fig.~\ref{fig:error_rates}) show false negatives are more likely in crowded scenes, which is not reflected in the summary detection metrics, since missing a certain amount of objects has stronger impact when less instances are present in the scene to begin with. Sparse images also have a substantially higher FP rate (0.19 vs 0.08), suggesting that performance gaps are mainly driven by increased background confusion. This can be attributed to the lack of context and large empty regions, increasing ambiguity. Meanwhile, crowded scenes with many co-occurring objects may offer stronger contextual cues and provide more positive training signals per image, reinforcing detections. Overall, layout represents an influential domain factor with clear performance gaps, highlighting the relevance of our scene composition axis beyond image appearance.

\textbf{Scale.} 
Scale has a significant and expected impact on performance. Large objects consistently outperform small objects across all metrics (Table~\ref{tab:condition_breakdown}). The gap is particularly pronounced in mAP50-95 and recall, indicating that bounding box localization is highly scale-dependent and that the performance gap is mainly driven by missed detections. This is coherent with the substantially higher FN rate (second highest across all domains) for small targets, while FP rates are more similar and even slightly higher for large objects. This suggests that the key issue is under-detection of small objects. This behavior is expected, as larger objects provide more pixels with stronger feature representations, and small objects contain limited visual information. Our scale category adequately captures this fundamental detection challenge.

\textbf{Background.} 
Background shows the strongest apparent performance difference, with complex scenes drastically outperforming simple ones by up to $\sim$16 points for mAP50 and recall (Table~\ref{tab:condition_breakdown}). However, this trend seems misleading when examining error rates (Fig.~\ref{fig:error_rates}). While simple scenes are more prone to FPs, there are only minor differences regarding FNs, with complex scenes even having a slightly higher error rate. Inspection of per-class performance and precision–recall curves (Fig.~\ref{fig:pr_curves} a and b) reveal that this discrepancy is largely caused by the scallop class. Due to the very small quantity of scallop samples (48 instances in all sparse test images), scallops reveal unstable behavior and strongly bias the mAP scores. The remaining performance differences for all other classes are much smaller, yet still in favor of complex scenes. Simple backgrounds are signal-poor with weak contrast and features, whereas complex backgrounds provide richer structure and context to learn patterns and define boundaries. Overall, background remains a highly informative domain factor, however the extreme performance gap is amplified by class imbalance and instability in rare classes.

\subsection{Axis 3: Acquisition Geometry}
\textbf{Orientation.} 
In comparison with other categories, orientation shows a weaker effect on performance, with upright images slightly outperforming rotated images (Table~\ref{tab:condition_breakdown}). The main difference appears in precision, where rotated images show a clear drop (up to $\sim$5 points), indicating increased false positives, which is confirmed by higher FP rates in the error analysis (Fig.~\ref{fig:error_rates}). Rotation introduces variability in object appearance and pose, making bounding box placement slightly less stable. Yet the bias is limited, considering that the training data is relatively balanced regarding orientation (Fig.~\ref{fig:train_domain_distr}), indicating it is a secondary factor in our framework.

\begin{figure}
    \centering
    \begin{subfigure}{0.37\columnwidth}
        \centering
        \includegraphics[width=\linewidth]{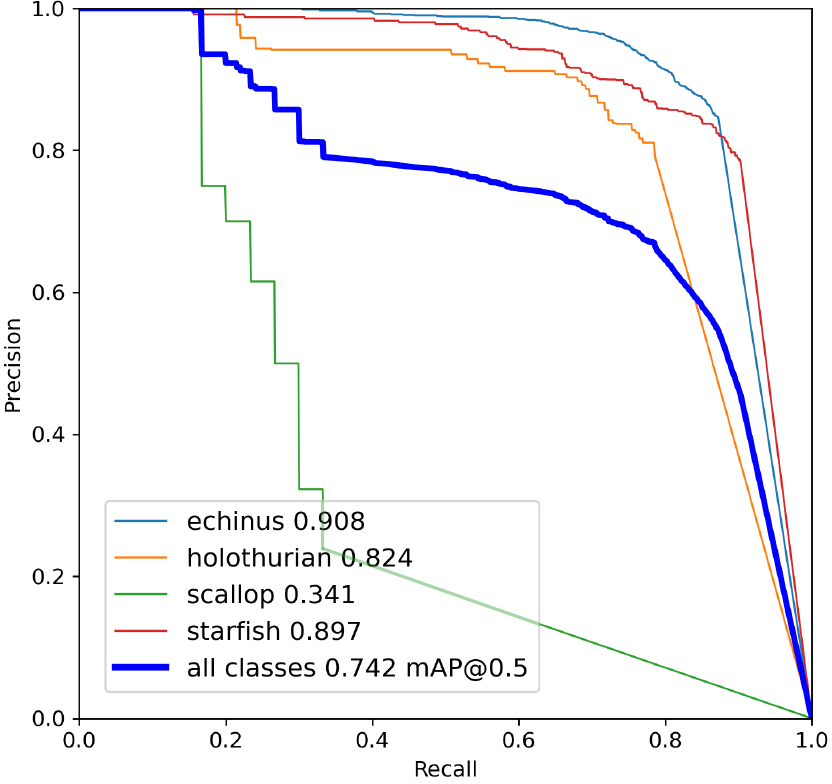}
        \caption{Simple}
    \end{subfigure}
    \hspace{0.05\columnwidth}
    \begin{subfigure}{0.37\columnwidth}
        \centering
        \includegraphics[width=\linewidth]{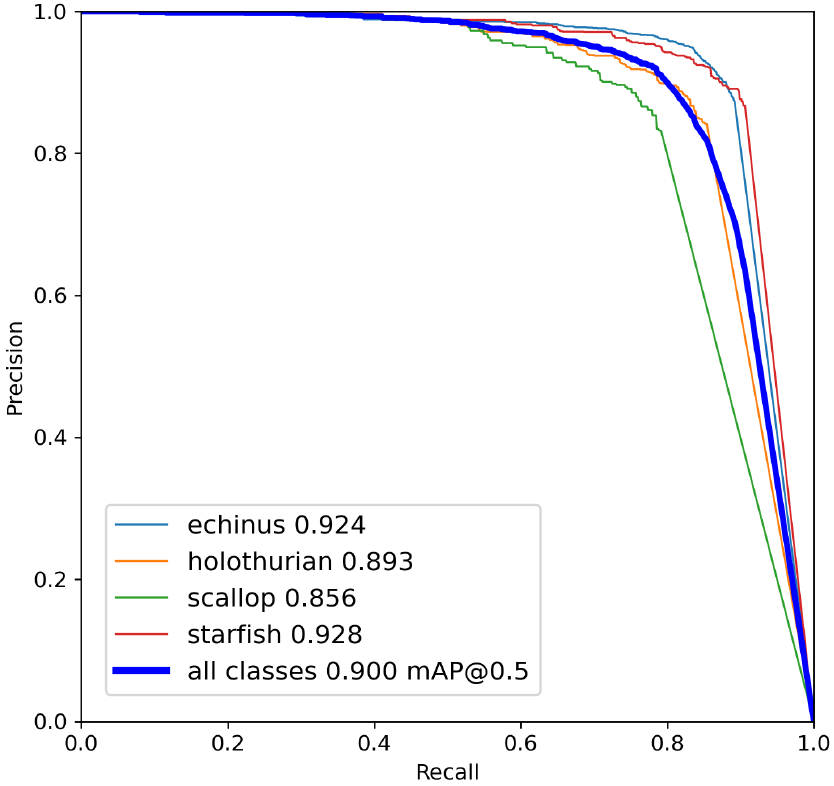}
        \caption{Complex}
    \end{subfigure}
    \vspace{0.5em}
    \begin{subfigure}{0.37\columnwidth}
        \centering
        \includegraphics[width=\linewidth]{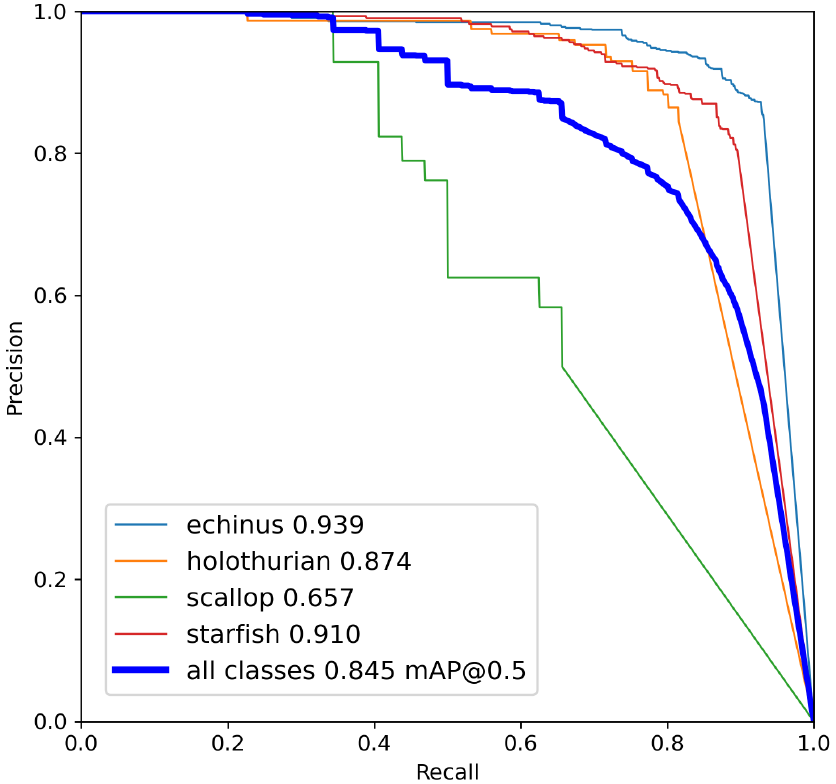}
        \caption{Nadir}
    \end{subfigure}
    \hspace{0.05\columnwidth}
    \begin{subfigure}{0.37\columnwidth}
        \centering
        \includegraphics[width=\linewidth]{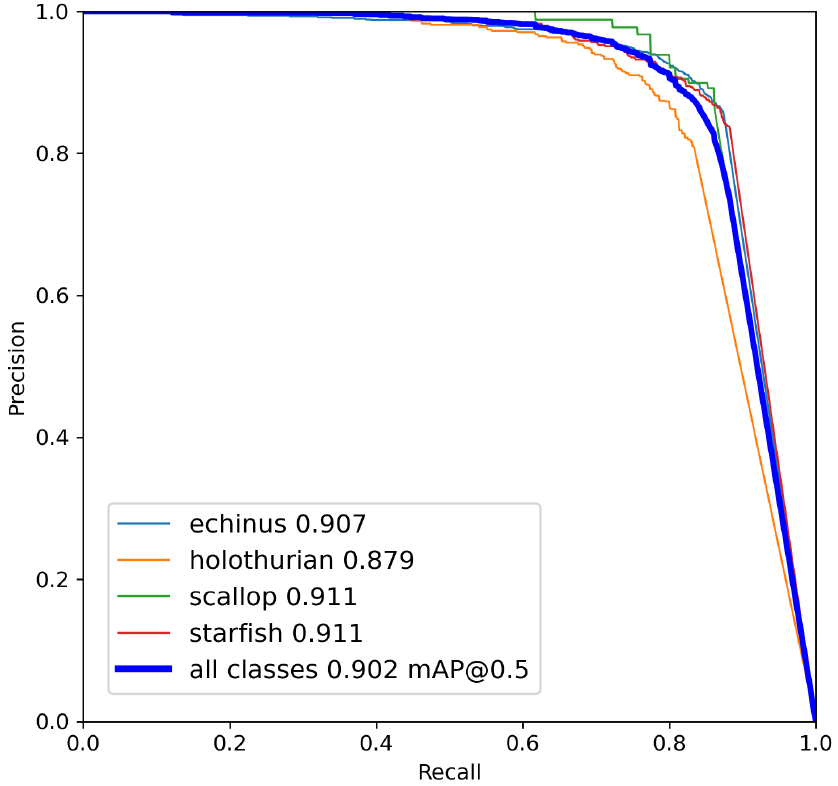}
        \caption{Front}
    \end{subfigure}
    \vspace{-0.3cm}
    \caption{Precision–recall (PR) curves for background (top) and perspective (bottom) categories. Scallops show stepwise curves with large jumps in simple (a) and nadir (c) subsets, caused by the very small number of ground-truth instances, disproportionately affecting the mAP score.}
    \vspace{-0.4cm}
    \label{fig:pr_curves}
\end{figure}

\textbf{Perspective.} 
Perspective has a complex influence on performance. Front views outperform nadir views for all metrics but most strongly for recall (Table~\ref{tab:condition_breakdown}), suggesting more missed objects in top-down views. However, raw error analysis (Fig.~\ref{fig:error_rates}) reveals a contradicting pattern: nadir images yield fewer FNs but more FPs. Similarly to background, this discrepancy is largely explained by class imbalance, as the nadir subset contains only 32 scallop instances (2.7\% of objects), which show significantly lower performance and introduce instability in the precision–recall curves (Fig.~\ref{fig:pr_curves}). For the remaining classes, performance in nadir is comparable or even slightly better. Simplified scenes but flattened target appearance in nadir views may benefit classes with distinctive silhouettes, while the depth information and more complex interactions in front views could favor structurally defined classes. Thus, perspective is a class-dependent domain factor. The observed performance gap is largely driven by rare, viewpoint-sensitive classes, highlighting the importance of domain-aware and class-aware evaluation.

\section{Conclusion and Outlook}
We studied underwater domain shift from a data-driven perspective, by separating meaningful domain factors (RQ1), and investigating their effect on detection performance (RQ2).

\textbf{RQ1:} We proposed a domain labeling framework that captures domain variations across image appearance, scene composition, and acquisition geometry, disentangling them into sets of interpretable factors (visibility, illumination, color; layout, scale, background; orientation, perspective). By quantifying the properties of each category using established metrics and ideas grounded in prior image quality assessment work, we could meaningfully group underwater imagery. The resulting pseudo domains aligned with semantic, visual, and physical conditions, reflecting real-world environments and representing domain variability as an explicit and structured variable.

\textbf{RQ2:} Our defined domain factors showed clear effects on detection performance. Visibility and scale were primary drivers, with other factors showing moderate but still meaningful effects. Orientation had minor influence under diverse training, while perspective showed class-dependent trends. In several cases, the observed effects were counter-intuitive (opposite to data distribution or human perception), highlighting the need for interpretable analysis of both performance trends and failure modes across domain properties in isolation.

This work highlights the importance of domain-aware evaluation and shows domain shift can be decomposed into interpretable factors. Our domain labeling framework enables transparent understanding of model behavior and reveals domain-specific limitations.

\textbf{Outlook.} Future work will extend this analysis in several directions. First, we observe dependencies between domain factors and object classes, but more systematic analysis under balanced data is needed. Second, domain labels can be used for domain-aware training strategies, such as targeted data augmentation, cross-domain training, or few-shot adaptation. Third, we are curating a large-scale domain-labeled dataset as a consistent benchmark to support domain generalization research for underwater object detection. 


\section*{Acknowledgment}
This research was supported by the QUT Centre for Robotics, QUT Digital Research Infrastructure team for HPC, and an ARC DECRA Fellowship DE240100149 to TF.

{\small
\bibliographystyle{ieee_fullname}
\bibliography{references}
}

\end{document}